\documentclass[11pt]{article}
\pdfoutput=1
\usepackage{naacl2021}

\usepackage{times}
\usepackage{latexsym}
\usepackage{graphicx}
\usepackage{ulem}

\usepackage[T1]{fontenc}

\usepackage[utf8]{inputenc}

\usepackage{microtype}

\usepackage{array}
\usepackage{alltt}
\usepackage{siunitx}
\usepackage{tabularx}
\usepackage{multirow}
\usepackage{todonotes}
\usepackage{listings}
\usepackage{verbatim}

\title{Recognizing and Splitting Conditional Sentences for Automation of Business Processes Management}

\author{Ngoc Phuoc An Vo, Irene Manotas, Octavian Popescu, Algimantas Cerniauskas, Vadim Sheinin \\
  IBM Research, Yorktown Heights, US \\
  \texttt{\{ngoc.phuoc.an.vo, irene.manotas, algimantas.cerniauskas\}@ibm.com} \\
  \texttt{\{o.popescu, vadims\}@us.ibm.com} \\}

\begin{document}
\maketitle
\begin{abstract}
Business Process Management (BPM) is the discipline which is responsible for management of discovering, analyzing, redesigning, monitoring, and controlling business processes. One of the most crucial tasks of BPM is discovering and modelling business processes from text documents. In this paper, we present our system that resolves an end-to-end problem consisting of 1) recognizing conditional sentences from technical documents, 2) finding boundaries to extract conditional and resultant clauses from each conditional sentence, and 3) categorizing resultant clause as Action or Consequence which later helps to generate new steps in our business process model automatically. We created a new dataset and three models solve this problem. Our best model achieved very promising results of \textit{83.82, 87.84}, and \textit{85.75} for Precision, Recall, and F1, respectively, for extracting Condition, Action, and Consequence clauses using Exact Match metric.

\end{abstract}

\section{Introduction}

Business processes which serve as basis for many (if not all) companies are usually stored as unstructured data, especially as text documents \cite{Blumberg03}. They can reflect all organization tasks and activities in order to provide services. Therefore, these documents could be converted into process models which allows to discover, analyze, redesign, monitor and control these business processes \cite{Dumas13}. The discipline which is responsible for management of this life cycle is known as Business Process Management (BPM).   
\par Today, this well known practice is very interdisciplinary and combines: organizational theory, management, and even computer science. One of the sub-areas of computer science is Natural Language Processing (NLP) and due to large number of freely availability NLP tools such as taggers, parsers, and the lexical database WordNet, NLP is applied in a multitude of domains and BMP is no exception. The application of NLP techniques could be used to extract useful information from textual documents to infer a process model, and also to extract information from the process model to facilitate visual process analysis \cite{Leopold13}. 
\par Process modelling is a very complex and time-consuming task. In this work we mainly focused on analysis of extracted information from text documents in order to have more accurate process models. Given a list of extracted sentences from technical documents, the contributions of our work consists of the followings: 1) classifying if a sentence is a conditional sentence, 2) identifying and extracting the conditional and resultant clauses from a conditional sentence, and 3) categorizing the extracted resultants.
Section 2 introduces business context and application for our tasks. Section 3 explains how we defined the tasks and created dataset towards solving the tasks. Section 4 presents our methods and models for the tasks. Section 5 shows experiments, evaluation results, and the error analysis that identifies the limitations of our models.

\footnotetext{\url{https://www.foodandbeveragetrainer.com/sop/}}

\section{Business Context, Application, and Related Works}

\paragraph{Process Discovery}

Traditional process models based on interviews often provide only a limited or biased picture of the actual process. Such highly abstract result can be interpreted in many ways. Therefore, during the past decade process mining and visualization tools (i.e., Celonis, UIPath) were developed to improve process visibility by generating highly adaptable, highly maintainable and validated business process models. However, these tools focus only on analysis of results from structured data - process sequences from actual user desktops, or systems logs. It is well known that Desktop Procedures (DTPs) and Standard Operation Procedures (SOP's) are very common process documentation techniques \cite{Phalp07}. Therefore, in order to see a broader picture of the process the information from textual source should be analyzed and included into process model. One of the tools which includes multiple sources is Process Discovery Accelerator (PDA). 

\begin{figure*}[!t]
    \centering
    \includegraphics[width=16cm]{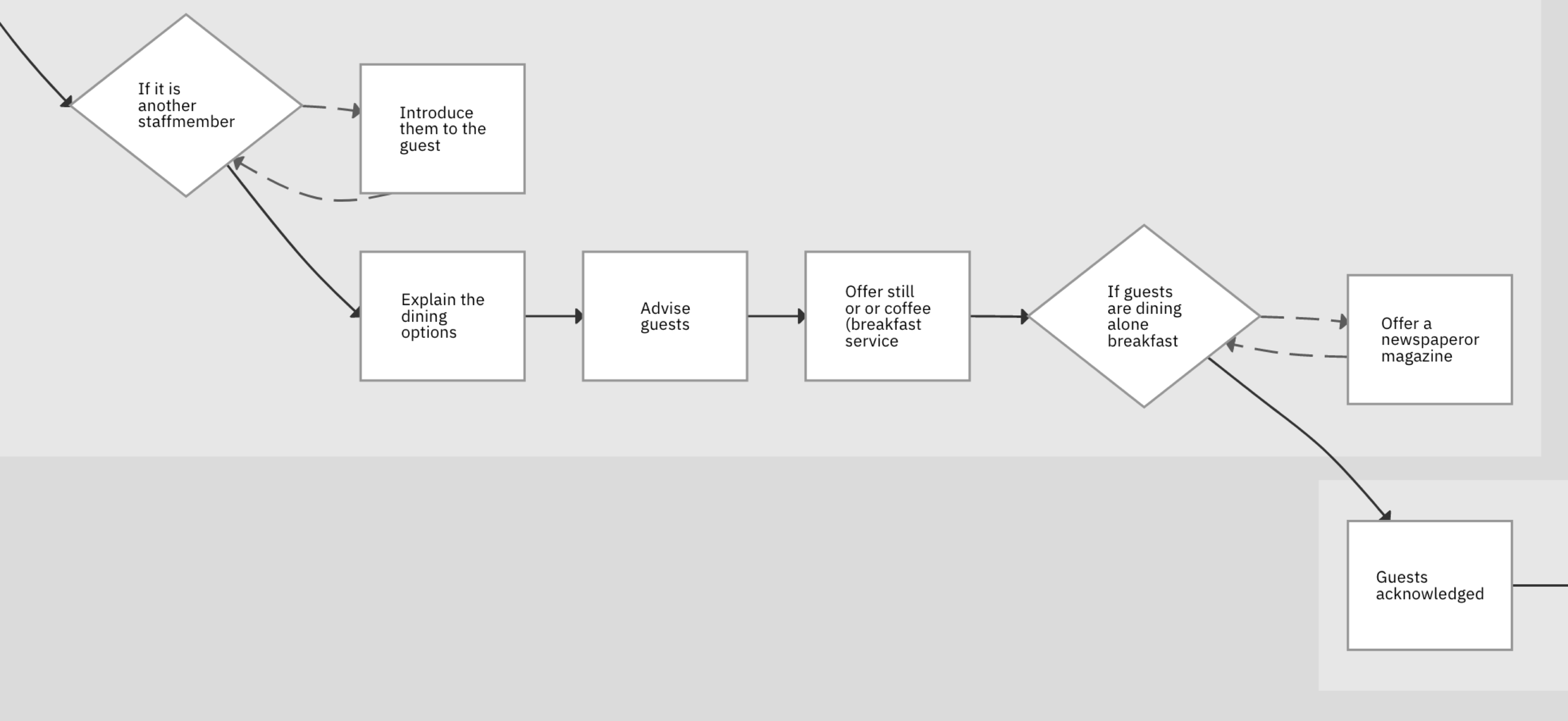}
    \caption{Standard operation procedure example of welcoming guest in restaurant from food and beverage training~\footnotemark~in Process Discovery Accelerator tool UI. The process step in diamond shape indicates condition clause and dashed arrow shows resultant. Solid lines connects different process steps.}
    \label{fig:PDA_example}
\end{figure*}

\paragraph{Application}
PDA is a business process mining and discovery tool currently available only for IBM automation consultants. It runs as a web application in production on the IBM Services Essentials for Automation platform. PDA has three main components: 1) SOP file analysis; 2) Click stream file analysis, sequential structured data which represents actual users activities; and 3) system logs file analysis which is also structured data from applications (system, e-mail, chat). PDA is able to discover processes using separate source components or all available at once, this procedure is called process harmonization. The final view of this procedure is single, correlated or uncorrelated processes from all available sources.
The example of a single (only SOP) simple process graph result in PDA UI is shown in Figure \ref{fig:PDA_example}. It illustrates how process flow looks like when conditional and resultant clauses are split into separate steps.
\par The SOP analysis is very challenging not only because it is based on unstructured data but also because it is very non-homogeneous, domain specific and biased since it is highly dependent on a process description creator \cite{barier}. These SOP files usually contains complex step descriptions, i.e., "if something, then do this" (Table \ref{tab:examples}) which sometimes could act as gateways or decision points. In order to have a more accurate process model we need to detect specific clause (conditions and actions) from already extracted process steps.
\par The main objective of PDA is to find automation opportunities by analyzing available data from SOP's, Click stream, and log files. Without accurate analysis of SOP files it is difficult to measure differences between different sources. Our proposed solution for conditional splitting is crucial for the SOP analysis, since it allows to see actual decision points and how complex a process is by detecting actions and/or consequences.

\paragraph{Related Works} \cite{Wenzina13} proposed heuristic, rule-based method to identify condition-action sentences. By using a set of linguistic patterns authors were able to split up sentences semantically into their clauses showing the condition and the consequence with recall of 75\% and precision of 88\%. Note: these numbers are for specific patterns - for all activities recall is 56\%.
Another rule-based work was presented by \cite{Hematialam17}. This work is based on combinations of part of speech tags used as features. Authors applied more statistical approach to automatically extract features, however, the recall value is very similar to \cite{Wenzina13}. Both works were tested using sentences from medical SOP's.
\par Another good example of NLP and BPM combination is the work from \cite{Qian2019ExtractingPG}. Authors used latest NLP techniques (transformers) to classify textual information from word-level to sentence-level. Although this work is most similar to our work it focuses more on automatic process graph extraction, which aims to extract multi-granularity information without manually defined procedural knowledge.

\section{Task Definition and Dataset Creation}
In this section, we first analyzed the sampled data from clients, then we defined the related task and created a new dataset for the task.

\subsection{Preliminary Data Analysis} We started with a set of 4,098 sampled sentences provided by our client from PDA SOP documents that were manually annotated in four categories with unbalanced label distribution:

\vspace*{-2mm}
\begin{itemize}
    \item No Condition (NC) (86.5\%): sentence that has no conditional logic, no condition is found.
    \vspace*{-2mm}
    \item Condition Action (CA) (9\%): sentence that has condition, and resultant is an Action.
    \vspace*{-2mm}
    \item Condition Consequence (CC) (3\%): sentence that has condition, resultant is a Consequence.
    \vspace*{-6mm}
    \item Only Condition (OC) (1.5\%): sentence that has only condition, no resultant found.
\end{itemize}

\begin{figure}
    \centering
    \includegraphics[scale=1.1]{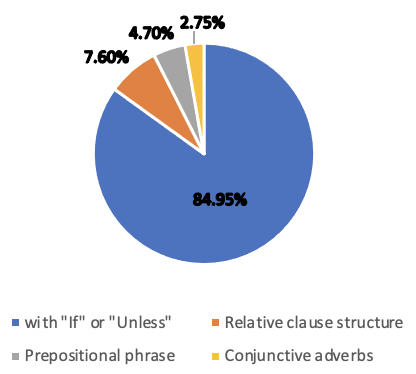}
    \caption{Syntactic structures of CA sentences.}
    \label{fig:CA_syntax}
\end{figure}

\begin{figure}
    \centering
    \includegraphics[scale=1]{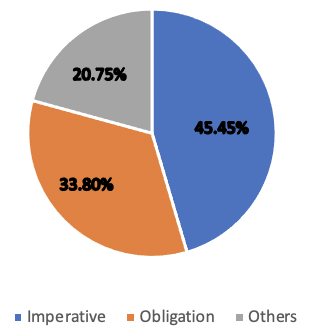}
    \caption{Action resultant types of CA sentences.}
    \label{fig:CA_resultant_types}
\end{figure}

\vspace*{-2mm}
We extracted 352 CA sentences then analyzed syntactic structures of CA sentences and different types of Action resultants (Figures \ref{fig:CA_syntax} and \ref{fig:CA_resultant_types}).

\subsection{Task Definition}
\label{sec:tasks}
We defined the end-to-end task Conditional Sentence Splitting in which the input is a given sentence, and the output are the extracted clauses classified into Condition and other resultant categories such as Action or Consequence. Since this is a large and complex task, it can be split into the following three subtasks.

\paragraph{Subtask 1: Conditional Sentence Classification}
This task is to recognize whether a given sentence is in conditional form. Conditional sentence is a sentence that expresses one or more thing(s) contingent on something else. The most basic form of a conditional sentence is \begin{equation} \label{eq:1} If P, Q \end{equation} where \textit{P} is the conditional clause, and \textit{Q} is the resultant. Recognizing conditional sentences is not a trivial task because there are various ways of expressing the meanings in conditional sentences. A basic approach to identify conditional sentences is to search for a conditional clause within a sentence that has keywords like \textit{If} or \textit{Unless} as condition indicators. However, conditional sentences do not always appear in standard conditional form in equation (\ref{eq:1}) with these indicators.
Table \ref{tab:examples} shows examples of conditional sentences that appear in different ways.

\begin{table*}[t]
\begin{center}
\begin{tabular}{l|c|c}
 Index & Conditional Sentences & Actual Meanings \\
\hline
 1 & If it rains, children should stay home. & If it rains, children should stay home. \\
 2 & Unless it rains, children can go out. & If it does not rain, children can go out. \\
 3 & 1. Otherwise, they can go out. & If it does not rain, children can go out. \\
 4 & Come now and I'll give you the book. & If you come now, I'll give you the book. \\
 5 & Do you like it? You can have it now. & If you like it, you can have it now. \\
 6 & For rainy days, children stay home. & If it rains, children stay home. \\
 7 & Anyone who skips class will be disciplined. & If anyone skips class, they will be disciplined. \\
\end{tabular}
\end{center}
\caption{Examples of Conditional Sentences appear in various ways.}
\label{tab:examples}
\end{table*}

\paragraph{Subtask 2: Sentence Boundary Finding and Splitting}
This task requires to find the boundary of conditional and resultant clauses in a conditional sentence. A basic approach is to locate the comma that separates between conditional and resultant clauses. For example: \textit{If it rains\textbf{,} children stay inside.} The comma in this example is a strong indicator showing that the first part of the sentence that has condition indicator \textit{If} is a conditional clause, and the second part is a resultant. However, as conditional sentences can appear in numerous ways (Table \ref{tab:examples}), this makes conditional sentence splitting a complex task.

\paragraph{Subtask 3: Resultant Categorization}
After splitting a conditional sentence into conditional and resultant clauses, the next step is to categorize if the resultant is an Action or a Consequence clause. In our application, we are interested in Action clause as it helps us to create a new action branch for our process model. For example:

\begin{itemize}
    \item \textit{Refer to the author if you are in any doubt about the currency of this document.} The resultant "\textit{Refer to the author}" is an Action clause which means the action will be taken should the condition be satisfied.
    
    \item \textit{If the entered password is matched with the one stored in system, the user is authenticated.} The resultant "\textit{the user is authenticated}" is a Consequence clause which shows the result should the condition be satisfied. In this case, no action is captured and no action branch is created for our process model.
\end{itemize}

For Action resultant (Figure \ref{fig:CA_resultant_types}), some common types we found are:

\vspace*{-3mm}
\begin{itemize}
    \item Imperative type is phrase that has an imperative verb to give a command or an order.
    \vspace*{-2mm}
    \item Obligation type is phrase that expresses something necessary to follow or execute. It usually has modals of obligation such as \textit{must, have to, have got to, need to, etc.}
    \vspace*{-2mm}
    \item Others are phrases that are neither in imperative nor in obligation form but actual meanings express the same.
\end{itemize}

\vspace*{-2mm}
Figure \ref{fig:PDA_example} shows an example where Condition Action sentences are transformed into process steps in our PDA system.

\subsection{Dataset Creation}
\label{sec:dataset}
The annotation task involved labeling contiguous sequences of words that are part of a conditional sentence i.e., the phrase or clause that define the condition and the resultant, respectively. In the annotation, is very important to detect correctly the extent of the annotation.

\paragraph{Data Source}
To extract examples of conditional sentences from real text documents, we collected public technical and regulations documents from different websites. Technical documents included manuals and troubleshooting guides for different software products and websites. Documents were processed by a parser to identify sentences. Using parser information about sentences features, we filter out sentences having conditional constructs. A total of \(1,936\) sentences with conditional statements were extracted for annotation.


\paragraph{Annotation Methodology}
We selected the Doccano~ \footnote{\url{https://doccano.herokuapp.com/}} annotation tool to annotate sequences in the sentences according to our task.
Guidelines for the annotation job included the possible types of sentences and clauses to annotate. The labels included in the annotation were the following:
\begin{itemize}
    \vspace*{-2mm}
    \item Condition (CD): Clause describing a condition. A conditional clause can come before or after the main clause.
    \vspace*{-2mm}
    \item Action (AC): Clause describing an actionable result that is related with a condition. The condition can be described before or after the Action clause in the same sentence.
    \vspace*{-2mm}
    \item Consequence (CS): Clause describing a non-actionable activity i.e., it does not describe an action that a system or user needs/should take.
    \vspace*{-6mm}
    \item Only-Condition (OC): Sentence describing only a condition without additional clauses and description of actions or activities.
    \vspace*{-2mm}
    \item No Condition (NC): Sentence with no conditional logic.
    \vspace*{-2mm}
    \item Unconditional-Action: Clause describing only an action without an associated condition.
\end{itemize}

\begin{center}
\begin{table}[t!]
\centering
 \begin{tabularx}{0.90\columnwidth}{@{}l|r|r|r|r|r|r@{}} 
Data & 	CD  	& 	CS 	& AC 	& OC & NC & UA \\ 
\hline
Train 	& 	1,446 & 	568 	& 886  	& 18 &  26 &  83\\
Test 		& 	184 	& 	71	& 114	&  3  &  4  & 13	\\
Dev 		&	184	&	67	& 114	&  3  &  5 & 23\\
Total 	& 	1,814 &	706 	& 1,114 	& 24 & 35 &  122\\
\end{tabularx}
\caption{Frequency of labels by data split.}
\label{tab:freq}
\end{table}
\end{center}

\vspace*{-7mm}
Annotators were guided to use labels at the sentence or clause level. \textit{Sentence-level} labels included No-Condition (NC) and Only-Condition (OC) labels, and can only be assigned to a whole sentence. \textit{Clause-level} labels included Condition (CD), Action (AC), and Consequence (CS) labels can be assigned only to a clause in a sentence having at least two clauses.

Before the annotation task, annotators were guided to analyze patterns of conditional sentences that could be found during the annotation task. For instance, sentences with short phrases such as \textit{if possible} or \textit{if any}, were not considered conditional phrases. 
Three annotators completed the annotation job for \(1,936\) sentences. Table \ref{tab:freq} shows the final number of annotated sequences per label in the  conditional label set.

\begin{table*}[t]
\begin{tabular}{l|c|c|c|c}
\hline
 & Rule-based & BERT-based & XLM-R-based & \\ 
\hline
 & \begin{tabular}{@{}llr@{}}
                   Precision & Recall & F1\\
                 \end{tabular}  
        & \begin{tabular}{@{}llr@{}}
                   Precision & Recall & F1\\
                 \end{tabular}  
        & \begin{tabular}{@{}llr@{}}
                   Precision & Recall & F1\\
                 \end{tabular} 
        & support \\
\hline
Condition & \begin{tabular}{@{}llr@{}}
                    80.52 & 76.86 & 78.65 \\
                 \end{tabular}  
        & \begin{tabular}{@{}llr@{}}
                    84.21 & 92.56 & \textbf{88.19} \\
                 \end{tabular}  
        & \begin{tabular}{@{}llr@{}}
                    78.12 & 81.52 & 79.79 \\
                 \end{tabular} 
        & 242 \\

Action & \begin{tabular}{@{}llr@{}}
                    69.74 & 70.2 & 69.97 \\
                 \end{tabular}  
        & \begin{tabular}{@{}llr@{}}
                    79.63 & 85.43 & 82.43 \\
                 \end{tabular}  
        & \begin{tabular}{@{}llr@{}}
                    80.24 & 88.74 & \textbf{84.28} \\
                 \end{tabular} 
        & 151 \\
        
Consequence & \begin{tabular}{@{}llr@{}}
                    51.52 & 55.43 & 53.4 \\
                 \end{tabular}  
        & \begin{tabular}{@{}llr@{}}
                    67.92 & 78.26 & 72.73 \\
                 \end{tabular}  
        & \begin{tabular}{@{}llr@{}}
                    88.21 & 89.67 & \textbf{88.93} \\
                 \end{tabular} 
        & 92 \\
\hline    
Average & \begin{tabular}{@{}llr@{}}
                    71.66 & 70.72 & 71.16 \\
                 \end{tabular}  
        & \begin{tabular}{@{}llr@{}}
                    79.7 & 87.63 & 83.46 \\
                 \end{tabular}  
        & \begin{tabular}{@{}llr@{}}
                    \textbf{83.82} & \textbf{87.84} & \textbf{85.75} \\
                 \end{tabular} 
        & 485 \\
\end{tabular}
\caption{Conditional Sentence Splitting performance of three models on Test set.}
\label{tab:results}
\end{table*}

\section{Model Descriptions}
We present three models using both rule-based and deep learning approaches to recognize and split conditional sentences into conditional clause and other resultant categories. The rule-based model implements a pipeline that solves the end-to-end task by solving each subtask described in (Section \ref{sec:tasks}). In contrast, we considered this large task as a sequential labeling task for deep learning approach. We used two state-of-the-art language models BERT and XLM-R that showed very good performance on sequential labelling task to train and fine-tune on our new dataset to solve the end-to-end task.

\subsection{Rule-Based Model}
The rule-based model uses three mechanisms to identify the conditional and resultant type in a sentence. The ESG parser \cite{McCord12} is used to identify the linguistic features using the dependency and constituency parser results. Then, we performed an alignment of the constituency and dependency parse results, to obtain features for every constituent and every word token. Finally, rules for conditional and resultant extraction are applied using the linguistic features available.
The output of the parser was mainly used to extract lexical and semantic features, like "imperative", for example, of the words in the sentence context. Our rules override the constituent borders. There are five categories of rules: (i) determining the scope of the conditional, (ii) determining the form of the predicate inside the conditional clause, (iii) determining the candidate clauses that could be linked to the conditional clause, (iv) determining resultant from candidates, and (v) labeling resultant accordingly to the guidelines. The rules have a large degree of linguistics generality, as they do not depend on the occurrence of overt lexical markers, or specific punctuation marks, e.g. comma etc.


\subsection{BERT-based Model}
One sequence model for conditional splitting was based on the ${BERT_{BASE}}$ language model~\cite{Devlin-etal'19-bert}. This model takes a maximum
512 input word piece token sequence \(X = [x_{1}; x_{2}; : : : ; x_{T}]\) 
and uses a \(L = 12\) layer transformer network, with 12 attention heads and 768 embedding dimensions, to output a sequence of contextualized token 
representations. We used the representation of the first sub-token as the input to the token-level classifier over the conditional label set. We fine-tuned the model using five epochs, with a learning-rate of \(0.1\), and using a \(256\) maximum sequence length.

\subsection{ XLM-R-based Model}
XLM-RoBERTa (XLM-R) is a state of the art multilingual masked language model trained on 2.5 TB of newly created clean CommonCrawl data in 100 languages \cite{conneau2019unsupervised}. It obtains strong gains over previous multilingual models like mBERT and XLM on classification, sequence labeling and question answering. We used the ${XLM-R}$ model with the following setting: max sequence length 256, learning rate 7e-5, warmup\_proportion 0.1, epoch 5, and batch size 8.

\section{Experiments}
We created a test set of 200 sentences from the dataset (Section \ref{sec:dataset}) following the data split (Table \ref{tab:freq}) for evaluating our three models.

Since our dataset was annotated at phrase/chunk level, we transformed our annotated data into IOB scheme for sequential labeling task. For example:

\vspace{-2mm}
\begin{itemize}
    \item Phrase level annotation: \textit{\{"id": 908, "text": "Include the date if the opt-out period expires.", "meta": {}, "annotation\_approver": "admin", "labels": [[0, 16, "Action"], [17, 47, "Condition"]]\}}
    \vspace{-1.5mm}
    \item IOB annotation scheme: \textit{\{Include B-Action, the I-Action, date I-Action, if B-Condition, the I-Condition, opt-out I-Condition, period I-Condition, expires I-Condition, . O\}}
\end{itemize}

\vspace{-2mm}
For our application, we are only interested in having sentences split into three main categories Condition, Action, and Consequence.
Table \ref{tab:results} shows the performance of our models for the end-to-end conditional sentence splitting on the test set of 200 sentences. The XLM-R-based model outperforms other two counterparts by a large margin.

\begin{figure}
    \centering
    \includegraphics[scale = 0.5]{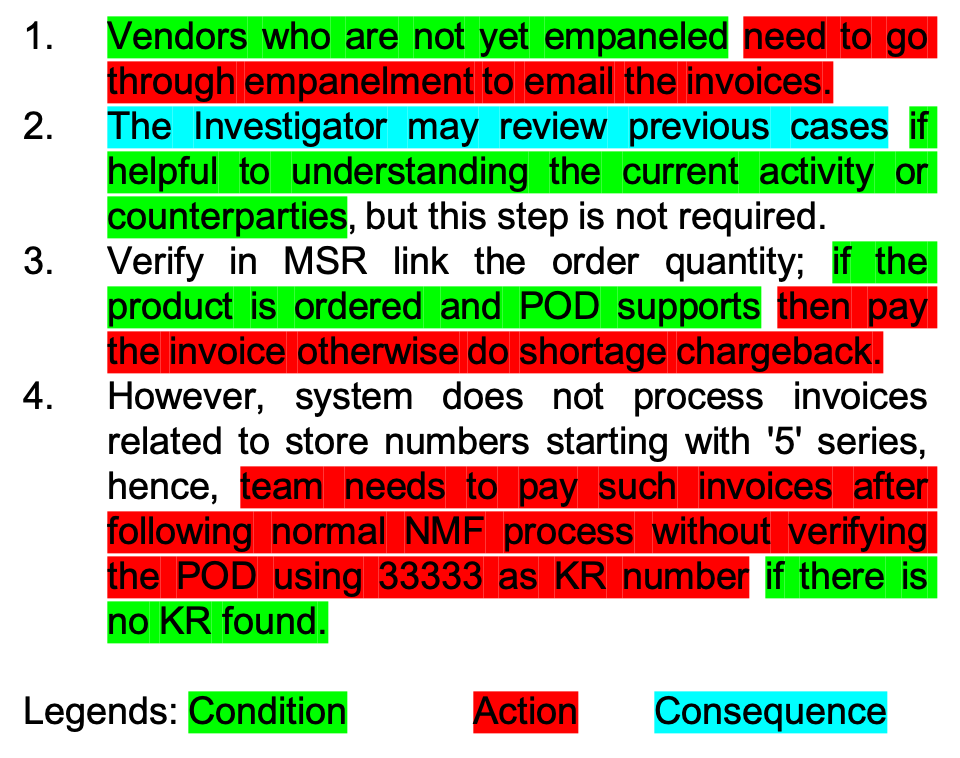}
    \caption{Sentences that our models failed to process. Annotation is shown for expected splitting and categorization.}
    \label{fig:error}
    \vspace{-2mm}
\end{figure}

\paragraph{Error Analysis} 
Figure \ref{fig:error} shows sentences having complex structures that our rule-based and deep learning models failed to recognize and extract correct Condition, Action, and Consequence clauses.

Although the rule-based model has a comprehensive set of rules for high accuracy, the coverage is limited for sentences having complex syntactic structures. It struggles to capture non-standard conditional structures, such as relative clause where no Condition clause was found (Example 1); or it captured wrong resultant (Example 2) due to structure complexity. In contrast, deep learning models performed well on (Examples 1 \& 2) but failed to capture correct resultant (Example 3) due to the coordination of verbs in the Condition clause; or failed to capture full resultant (Example 4) due to a very long sentence sequence. Our models also have difficulty to process sentences with multiple conditions or multiple resultants. For example: 

\vspace{-2mm}
\begin{itemize}
    \item \textit{\textbf{If you had dynamic SQL} (or \textbf{if you rebound static SQL}), your applications might be breached.}
    \vspace{-2mm}
    \item \textit{If using PayPal for payment \textbf{then click on the PayPal tab} and \textbf{then click Pay Now.}}
\end{itemize}

\vspace{-2mm}
Example 3 shows another limitation that is to differentiate between Unconditional-Action clause \textit{"Verify in MSR link the order quantity"} and the actual Action clause.

\section{Conclusions and Future Work}
Recognizing and splitting conditional sentences for automation of business process management is a complex and important task for many industries. In this paper we presented a non-trivial real-world system that can recognize and split technical instructions (at a sentence level) into Condition, Action, Consequence clauses for business process modeling and automatic update. We defined an end-to-end task that was split into three subtasks to fit our business context and needs. We also created a new dataset and three models, using both rule-based and deep learning approaches, to solve this task. Although there are limitations, our models have served our needs successfully. For future work, we plan to expand our dataset with more fine-grained annotation for complex sentences and improve the performance of our models.

\bibliography{naacl2021}
\bibliographystyle{acl_natbib}

\end{document}